\documentclass[lettersize,journal]{IEEEtran}
\usepackage[caption=false,font=normalsize,labelfont=sf,textfont=sf]{subfig}
\usepackage{amsmath,amsfonts}
\usepackage{array}
\usepackage{textcomp}
\usepackage{stfloats}
\usepackage{url}
\usepackage{verbatim}
\usepackage{graphicx}
\usepackage{cite}
\usepackage{ifpdf}
\usepackage[ruled,vlined]{algorithm2e}
\usepackage{amssymb}
\usepackage{booktabs} 
\usepackage{bbding}
\usepackage{pifont}
\usepackage{multirow}
\usepackage{booktabs} 
\usepackage{tcolorbox}
\usepackage[colorlinks=true,
            linkcolor=blue,
            anchorcolor=blue,
            citecolor=blue]{hyperref}
\definecolor{targetgreen}{RGB}{180,220,160}
\definecolor{sourceyellow}{RGB}{255,235,180}
\definecolor{outputorange}{RGB}{245,200,160}

\begin{document}

\title{Source-Free Domain Adaptation for Geospatial Point Cloud Semantic Segmentation}

\author{Yuan~Gao, Di~Cao, Xiaohuan~Xi, Sheng~Nie, Shaobo~Xia, Cheng~Wang

\thanks{This work was supported by the National Natural Science Foundation of China under Grant 62575039 and 42271365.
(\textit{Corresponding author: Shaobo Xia.})}
\thanks{Yuan Gao, Xiaohuan Xi, and Sheng Nie are with the Aerospace Information Research Institute, Chinese Academy of Sciences, Beijing 100094, China, and also with International Research Center of Big Data for Sustainable Development Goals, Beijing 100094, China, and University of Chinese Academy of Sciences, Beijing 100049, China. \par

Di Cao is with the Zhengzhou Institute for Advanced Research of Henan Polytechnic University, Henan Polytechnic University, Zhengzhou 451464, China.\par

Shaobo Xia is with the School of Aeronautic Engineering, Changsha University of Science and Technology, Hunan 410004, China (e-mail: shaoboxia2020@gmail.com). \par

Cheng Wang is with China University of Geosciences, Beijing 100083, China.
}
}

\maketitle

\begin{abstract}
Semantic segmentation of 3D geospatial point clouds is fundamental to remote sensing applications, yet domain shifts caused by regional and acquisition-related variations often degrade model performance. Although domain adaptation can mitigate such shifts, existing methods typically require access to source-domain data, which is often infeasible due to privacy concerns and regulatory policies. To address this, we propose LoGo (Local–Global dual-consensus), a novel source-free unsupervised domain adaptation (SFUDA) framework requiring only a pretrained model and unlabeled target data. At the local level, we introduce a class-balanced prototype estimation module that ensures that robust feature prototypes can be generated even for sample-scarce tail classes, effectively mitigating the feature collapse caused by long-tailed distributions. At the global level, we introduce an optimal transport-based global distribution alignment module that formulates pseudo-label assignment as a global optimization problem, effectively correcting the over-dominance of head classes inherent in local greedy assignments, and thereby preventing model predictions from being severely biased towards majority classes. Finally, we propose a dual-consistency pseudo-label filtering mechanism that retains only high-confidence pseudo-labels where local multi-augmented ensemble predictions align with global optimal transport assignments for self-training. Extensive experiments on two challenging benchmarks, encompassing cross-scene and cross-sensor settings, demonstrate that LoGo consistently outperforms existing state-of-the-art methods. The source code is available at \url{https://github.com/GYproject/LoGo-SFUDA}.

\end{abstract}


\section{Introduction}

\IEEEPARstart{T}{hree} dimensional geospatial point clouds, acquired via Airborne Laser Scanning (ALS), Mobile Laser Scanning (MLS), and photogrammetry, provide precise geometric representations of the real world \cite{11422025,CHEN2026675,WANG2025422}. They form the foundation for cadastral surveying \cite{xia2021building}, land cover mapping \cite{CHEN202518}, and 3D reconstruction \cite{10414143}, making semantic segmentation of large-scale 3D urban scenes a critical research topic. Despite the strong performance of deep learning-based methods for point cloud interpretation \cite{9127813}, their generalization to unseen large-scale geospatial scenes remains limited by domain shift~\cite{10818527,WANG2025422}. In geospatial data, such domain shift mainly arises from two sources: cross-scene variations and cross-sensor discrepancies. Cross-scene shift is caused by the intrinsic diversity of urban environments, where differences in architectural styles, city layouts, vegetation types, and object appearances lead to distributional discrepancies across regions. Cross-sensor shift, by contrast, stems from sensor-specific scanning geometry and sampling mechanisms, which induce variations in point density and spatial distribution patterns~\cite{WANG2025422,GAO2026339}.

Unsupervised Domain Adaptation (UDA) has emerged as a mainstream paradigm for mitigating domain shift~\cite{LUO2020253}. Conventional UDA assumes simultaneous access to labeled source data and unlabeled target data, and reduces cross-domain discrepancies through adversarial learning, feature alignment, or related strategies. While this setting avoids costly target-domain annotations, it is often impractical in surveying and remote sensing applications, where source data may be inaccessible due to national security, commercial sensitivity, or privacy regulations. In such cases, end-users usually have access only to a pre-trained source model. To address both domain shift and source-data inaccessibility, Source-Free Domain Adaptation (SFUDA) has emerged as a promising alternative. SFUDA adapts the pre-trained source model to an unlabeled target domain without accessing source data, making it particularly suitable for privacy-sensitive geospatial semantic segmentation~\cite{GAO202672}.

While SFUDA has shown promise, existing studies on 3D point clouds have predominantly focused on autonomous driving scenarios~\cite{michele2024train}, where domain shifts are usually limited to homogeneous LiDAR sensors with different beam counts and relatively small viewpoint variations. In contrast, SFUDA for heterogeneous geospatial point clouds remains underexplored and faces two major challenges. First, geospatial data exhibit substantial cross-modality and viewpoint discrepancies. Adaptation from photogrammetry-derived point clouds to ALS data involves fundamental structural differences between surface-only reconstruction and volumetric sensing, while shifts from airborne nadir-view systems to mobile ground-view systems can invalidate many geometry-dependent priors. Second, geospatial scenes suffer from pronounced class imbalance. For example, in the Toronto-3D dataset~\cite{tan2020toronto}, Road and Building account for approximately 77.83\% of all point cloud data, whereas Utility line and Fence constitute only 0.85\% and 0.52\%, respectively.  This long-tailed distribution makes tail classes prone to being overlooked, resulting in missed detections.

To bridge this gap, we propose LoGo, a novel Source-Free Domain Adaptation framework based on Local-Global Dual Consensus. LoGo freezes the pre-trained feature extractor and classifier, and updates only the Batch Normalization (BN) layers to adapt target-domain statistics while preserving source-domain geometric representations. For pseudo-label generation, LoGo introduces Class-Balanced Local Prototype Estimation, which mines high-confidence intra-class anchors to build robust prototypes and mitigate class imbalance. It then incorporates global distribution constraints via Optimal Transport (OT), where Sinkhorn-based alignment corrects the bias of local assignments using global class priors. Finally, Local-Global Dual-Consensus Filtering retains only pseudo-labels consistently supported by both local ensemble predictions and global OT assignments, thereby suppressing noise propagation under domain shift. Our main contributions are summarized as follows:
\begin{itemize}
\item We formulate source-free domain adaptation for geospatial point cloud semantic segmentation under severe cross-scene and cross-sensor shifts, including Photogrammetry-to-LiDAR adaptation, without accessing source data.

\item We propose LoGo, a Local-Global Dual-Consensus framework that combines class-balanced local prototype estimation with Optimal-Transport-based global distribution alignment to reduce long-tailed class bias and pseudo-label noise.

\item We conduct extensive experiments on two benchmarks spanning Photogrammetry, ALS, and MLS data, where LoGo achieves state-of-the-art performance and demonstrates strong potential for privacy-preserving large-scale geospatial mapping.
\end{itemize}

The remainder of this article is organized as follows. Section \ref{rework} reviews related work. Section \ref{method} introduces the proposed LoGo framework. Section \ref{experiment} presents experiments and ablation studies. Section \ref{conclus} concludes the article and discusses future directions.

\section{Related Work}\label{rework}

\subsection{Domain Adaptation for Point clouds}
3D point cloud semantic segmentation assigns semantic labels to individual points in unordered point sets~\cite{hu2020randla,thomas2019kpconv}. As fully supervised methods rely on costly point-wise annotations and generalize poorly under domain shift, UDA aligns unlabeled target data with the source domain to improve cross-domain performance~\cite{11521044}, with existing point cloud UDA methods mainly following adversarial feature alignment or self-training paradigms.

Adversarial training attempts to learn domain-invariant representations by minimizing the feature space distance between domains \cite{zhao2021epointda,yi2021complete,xiao2022transfer,li2023adversarially,yuan2023prototype}. ePointDA \cite{zhao2021epointda} proposes a self-supervised strategy simulating real-world degradation via Dropout noise on synthetic data. It combines statistics-invariant Instance Normalization with spatially adaptive convolution for feature alignment, thereby enhancing cross-domain segmentation performance without relying on real-domain statistical priors. Addressing differences in sensor characteristics, Li et al. \cite{li2023adversarially} observe that domain shift largely stems from the structural differences between complete synthetic point clouds and real LiDAR data, which contain irregular noise such as missing points. They employ an adversarial training mechanism to guide a masking module, which simulates real noise distributions, effectively reducing domain discrepancy through feature alignment.

Self-training strategies focus on iterative optimization using target domain pseudo-labels \cite{Kong2021ConDAUD,Luo_2021_ICCV,saltori2022cosmix,xiao2022polarmix,shaban2023lidar}. To preserve spatial semantic continuity, ConDA \cite{Kong2021ConDAUD} leverages the spatial structure of Range View (RV) images by constructing an intermediate domain through non-overlapping splicing of source and target image strips. This approach facilitates inter-domain interaction while maintaining semantic coherence around autonomous vehicles. Additionally, an entropy aggregator filters high-confidence pseudo-labels. LiDAR-UDA \cite{shaban2023lidar} focuses on mitigating domain shift caused by sensor configuration differences, particularly beam density and scanning patterns. This method employs structured point cloud downsampling to simulate diverse scan patterns, enhancing robustness to sparse data. Furthermore, by exploiting the temporal consistency of point cloud sequences to perform weighted fusion of multi-frame predictions, it generates high-quality refined pseudo-labels to guide self-training in the target domain.

To address complex cross-scene and cross-sensor challenges in geospatial point clouds, several studies propose targeted solutions. To address domain shift in urban MLS caused by varying urban morphologies (e.g., building height), Luo et al. \cite{LUO2020253} design a Point-wise Attention Transformation Module (PW-ATM) to align vertical height distributions in the input space. Subsequently, combined with Maximum Classifier Discrepancy (MCD), it aligns feature distributions in the latent space. Additionally, Luo et al. \cite{luo2025cross} utilize approximate unsigned distance fields to reconstruct continuous latent scene surfaces, mapping both source and target point clouds into a unified Canonical Domain.

\subsection{Source-Free Unsupervised Domain Adaptation}

Although effective, conventional UDA requires simultaneous access to source and target data, which is often impractical in surveying and remote sensing due to privacy, bandwidth, and storage constraints; this motivates Source-Free Domain Adaptation (SFUDA), which adapts a pre-trained source model to unlabeled target data without accessing source data and has been explored mainly in the 2D image domain through generative, fine-tuning-based, and self-training methods.

Generative methods \cite{kundu2020universal,kurmi2021domain,ijcai2021p402} synthesize pseudo-samples using Generative Adversarial Networks (GANs) or source model statistics to assist alignment \cite{9157645,9578923}, though often with high computational overhead. Fine-tuning-based methods focus on immediate inference-time adaptation. For instance, Tent \cite{wang2021tent} and DUA \cite{mirza2022norm} minimize prediction entropy or update Batch Normalization (BN) statistics online. 
In contrast, pseudo-label-based self-training methods \cite{yang2021generalized,yang2022attracting,tanwisuth2021prototype,karim2023c,10657922} typically adopt an iterative mode. 
SHOT \cite{liang2020we} freezes the classifier and maximizes information flow, employing a global prototype-based strategy to align target features with the source hypothesis. 
DT-ST \cite{10203421} stabilizes the direction of model evolution by introducing a dynamic teacher update mechanism, and mitigates the class bias of the source model by integrating a training-consistency-based resampling strategy for reliable samples.
SeCoV2 \cite{11120362} leverages vision foundation models to aggregate pseudo-labels into semantically consistent regions, and constructs a Graph Variational Autoencoder  to perform region-level uncertainty estimation and correction.

Recently, SFUDA has been extended to the 3D point cloud domain, primarily focusing on autonomous driving. Michele et al. \cite{michele2024train} propose a framework utilizing source class distribution priors and consistency with a reference model as unsupervised stopping criteria, achieving stable fine-tuning. TGSF \cite{duan2024source} employs a Bi-directional Pseudo-Label Selection, dynamically distinguishing between source-similar and source-distinct data based on entropy to extract reliable pseudo-labels, while leveraging a teacher-student architecture to enforce global and class prototype consistency to prevent model collapse.
Beyond segmentation, SFUDA has also been investigated in 3D object detection \cite{saltori2020sf,hegde2023source,10633799}, point cloud completion \cite{xia2025dspf,he2025pointsfda}, and primitive segmentation \cite{wang2024multi,11184737}.

For geospatial point clouds, source-free adaptation remains at an early stage. Liu et al.~\cite{11091594} address urban-scale segmentation by projecting 3D point clouds into BEV images and leveraging 2D foundation models with geometric-prior correction for pseudo-label generation. Wang et al.~\cite{WANG2025422} further explore test-time adaptation for 3D point clouds through reliability-guided progressive BN updates. However, these methods remain limited for heterogeneous geospatial transfer: BEV-based projection may lose fine-grained 3D geometric structures, while online batch-wise adaptation is sensitive to spatial heterogeneity and class imbalance. To overcome these limitations, we propose a 3D offline adaptation framework that preserves full geometric information and incorporates Optimal-Transport-based global optimization for robust distribution alignment under severe class imbalance.

\begin{figure*}[t]
    \centering
    \includegraphics[scale=0.17]{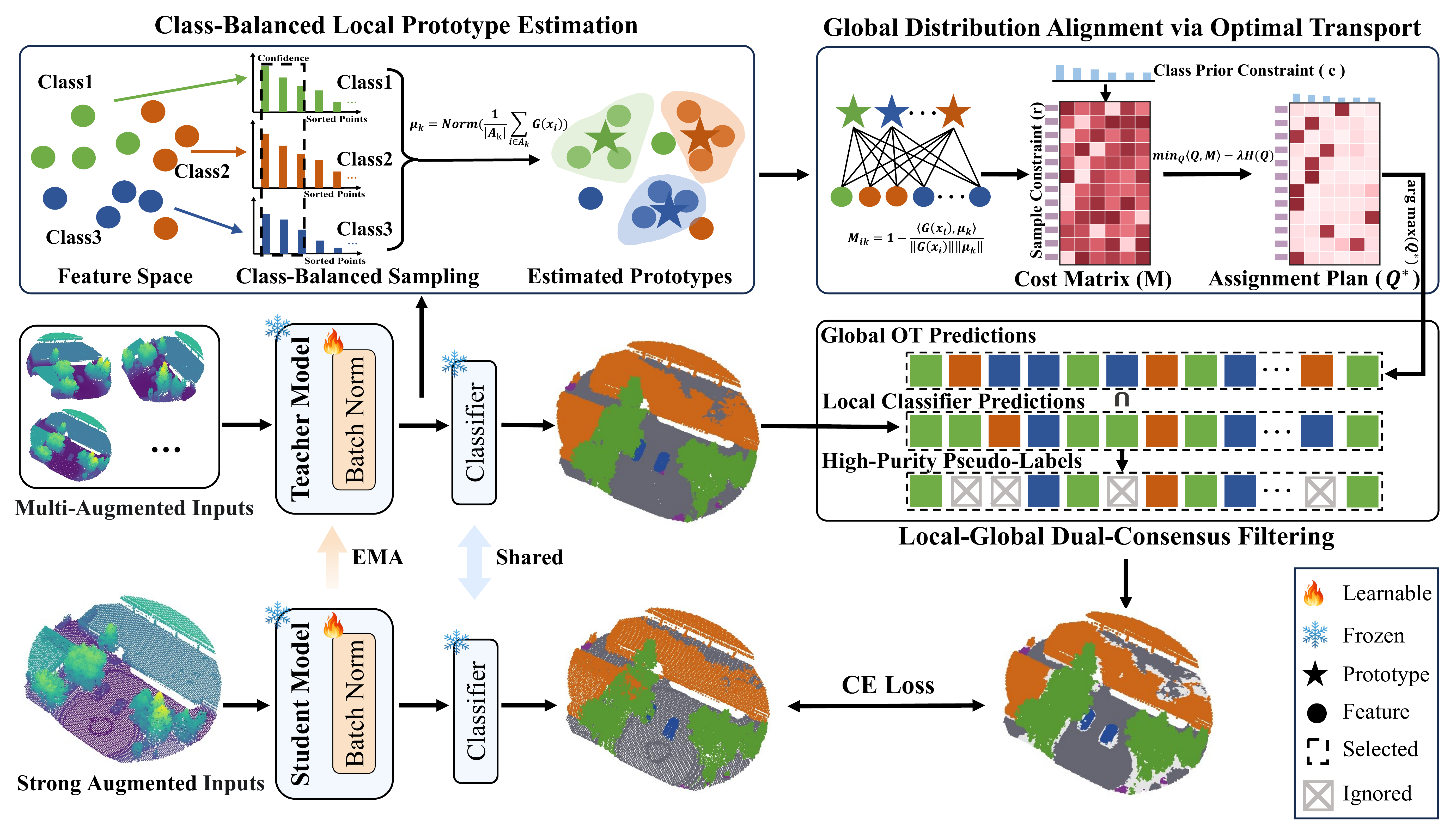} 
    \caption{Overview of the proposed LoGo framework. The architecture adheres to a parameter-efficient Mean-Teacher paradigm, where only the BN layers are learnable (indicated by the flame icon), while other parameters remain frozen (snowflake icon). The adaptation process begins by aggregating features from multi-augmented inputs via the Class-Balanced Local Prototype Estimation (CBLPE) module, which employs an intra-class anchor mining strategy to estimate robust class prototypes. Subsequently, the Global Distribution Alignment (GDA) module solves an Optimal Transport problem to generate a global assignment plan $\mathbf{Q}^*$ by minimizing transport costs under class prior constraints. Finally, the Local-Global Dual-Consensus Filtering (LGDCF) mechanism selects reliable pseudo-labels by identifying the intersection ($\cap$) between local classifier predictions and global OT assignments. The student model is then supervised by these refined labels on strongly augmented data via Cross-Entropy (CE) loss, while the teacher model evolves through Exponential Moving Average (EMA).}
    \label{fig:framework}
\end{figure*}

\section{Methodology}
\label{method}

To mitigate domain distribution shift and the lack of supervision in the SFUDA setting, we propose a robust self-training framework, termed LoGo (Local-Global Dual-Consensus). The overall pipeline is illustrated in Fig. \ref{fig:framework}.
We adopt an iterative ``Offline Self-Training” paradigm.
Built upon a parameter-efficient Mean-Teacher paradigm, our approach refines pseudo-labels through three core modules. 
First, to reduce prediction uncertainty, we employ a Multi-Augmented Ensemble strategy.  
Subsequently, the Class-Balanced Local Prototype Estimation (CBLPE) module constructs reliable feature prototypes from a local intra-class perspective. 
Following this, the Global Distribution Alignment (GDA) module leverages Optimal Transport to enforce global distributional constraints and regularize class assignments. 
Ultimately, the Local-Global Dual-Consensus Filtering (LGDCF) module integrates these distinct views to curate reliable pseudo-labels for student model supervision.

\subsection{Problem Formulation and Framework}

\subsubsection{Source-Free Unsupervised Domain Adaptation Setting} 

We formulate the SFUDA task within the context of 3D point cloud semantic segmentation. Formally, let $\mathcal{X}$ denote the input space comprising point coordinates and features, and $\mathcal{Y} = \{1, \dots, K\}$ represent the shared discrete label space. The segmentation network is decomposed into a feature encoder $G: \mathcal{X} \rightarrow \mathcal{F}$, which maps inputs to a high-dimensional feature space $\mathcal{F} \subseteq \mathbb{R}^D$, and a classifier $C: \mathcal{F} \rightarrow \mathbb{R}^K$, which projects features to class logits.

We consider a source domain $\mathcal{D}_S$ and an unlabeled target domain $\mathcal{D}_T = \{x_i\}_{i=1}^{N}$, where $N$ denotes the total number of target points.
Crucially, we operate under the Closed-Set assumption ($\mathcal{Y}_S = \mathcal{Y}_T$), tackling the challenge where feature distributions differ significantly ($\mathcal{P}_S(x) \neq \mathcal{P}_T(x)$).

The SFUDA protocol proceeds in two stages: first, the source model $M_S = C \circ G$ is pre-trained on $\mathcal{D}_S$ via standard supervision. Subsequently, in the adaptation phase, access to $\mathcal{D}_S$ is restricted. Our objective is to adapt the model parameters, initialized with pre-trained weights $\theta_S$, to align the target representation utilizing only the unlabeled target data $\mathcal{D}_T$.

\subsubsection{Parameter-Efficient Mean-Teacher Architecture}
To achieve robust adaptation under unsupervised conditions, we adopt a self-training framework based on the Mean-Teacher paradigm \cite{tarvainen2017mean}. This framework comprises a student model $M_{stu}$, updated via gradient descent, and a teacher model $M_{tea}$, evolved via Exponential Moving Average (EMA), as shown in Fig. \ref{fig:framework}.
Since full fine-tuning risks overfitting to noisy pseudo-labels and distorting the pre-trained feature space, we adopt a parameter-efficient strategy \cite{liang2020we}. Specifically, we freeze the convolutional weights of the feature extractor $G$ and the parameters of the classification head $C$, optimizing only the affine parameters of the Batch Normalization (BN) layers. This strategy enables the model to rapidly align with the target domain distribution by recalibrating feature statistics while preserving the discriminative power of the pre-trained representations.

\subsection{Class-Balanced Local Prototype Estimation}

Prior to each training epoch, we perform a global inference pass to update pseudo-labels. Without source supervision, reliable feature alignment depends on robust class prototypes. However, directly aggregating features from raw predictions is unreliable in geospatial point clouds with severe domain shifts and long-tailed distributions, as head-class prediction bias can contaminate tail-class prototypes and cause representation collapse. To address this issue, we propose the Class-Balanced Local Prototype Estimation (CBLPE) module, as shown in Fig. \ref{fig:framework}. Instead of using a global confidence threshold, which may suppress naturally low-confidence tail classes, CBLPE adopts a class-aware local selection strategy. By mining high-confidence anchors within each predicted category, it constructs distinct prototypes for under-represented classes while reducing contamination from dominant classes.

\subsubsection{Multi-Augmented Ensemble Inference}
To suppress prediction variance caused by point cloud sparsity, we introduce a multi-augmented ensemble mechanism.
At each epoch, the teacher model $M_{tea}$ processes the entire dataset $\mathcal{D}_T$ under $V$ augmentation transformations (e.g., rotation). We aggregate the predictions to obtain a smoothed probability distribution $\bar{\mathbf{p}}_i$:
\begin{equation}
\bar{\mathbf{p}}_i = \frac{1}{V} \sum_{v=1}^{V} \sigma(C_{tea}(G_{tea}(\mathcal{T}_v(x_i))))
\end{equation}
where $\mathcal{T}_v(\cdot)$ denotes the $v$-th random transformation, and $\sigma$ is the Softmax function. This process yields probability estimates that are more robust against input noise. 
Crucially, to ensure computational efficiency, this multi-augmented inference is executed entirely offline and in a gradient-free manner prior to each training epoch.

\subsubsection{Local Confidence Metric}
Based on the smoothed probability distribution, we extract the raw prediction label $\hat{y}_{raw, i}$ and confidence score $s_i$ for each sample:
\begin{equation}
\hat{y}_{raw, i} = \arg\max (\bar{\mathbf{p}}_i), \quad s_i = \max (\bar{\mathbf{p}}_i)
\end{equation}
where $s_i$ quantifies the model's certainty regarding the sample assignment.

\subsubsection{Intra-class Anchor Mining}
Traditional pseudo-label filtering strategies often rely on a global unified confidence threshold (e.g., $\delta=0.9$). However, under severe class imbalance, the prediction confidence for tail classes is consistently low. Consequently, these classes are prone to being filtered out, leading to a bias towards majority classes.

To address this, we propose an intra-class independent mining strategy that performs filtering separately within each category. For any class $k \in \{1, \dots, K\}$, we construct a candidate set $\mathcal{I}_k = \{i \in \{1, \dots, N\} \mid \hat{y}_{raw, i} = k\}$. Instead of a fixed threshold,  we select the top $\rho$ percentage of samples (e.g., $\rho=0.8$) to form the high-confidence anchor set $\mathcal{A}_k$:
\begin{equation} 
\mathcal{A}_k = \{ i \in \mathcal{I}_k \mid \text{rank}_{\mathcal{I}_k}(s_i) \le \lfloor \rho \cdot |\mathcal{I}_k| \rfloor \}
\end{equation}
\noindent 
where $\text{rank}_{\mathcal{I}_k}(s_i)$ denotes the ranking index of sample $i$ within the global candidate set $\mathcal{I}_k$ sorted by confidence in descending order.
This design follows the principle of relative reliability: it ensures that even for difficult tail classes, the most reliable subset is utilized to represent the class distribution.

\subsubsection{Feature Prototype Aggregation}
Based on the anchor set $\mathcal{A}_k$, the prototype $\boldsymbol{\mu}_k$ for category $k$ is calculated as the normalized mean of the features of the samples in the set:
\begin{equation}
\boldsymbol{\mu}_k = \text{Normalize}\left( \frac{1}{|\mathcal{A}_k|} \sum_{i \in \mathcal{A}_k} G_{tea}(x_i) \right)
\end{equation}
These prototypes serve as geometric centroids within the target feature space, representing the unique local structural characteristics of each category.

\subsection{Global Distribution Alignment via Optimal Transport}

With the constructed class-balanced prototypes $\boldsymbol{\mu}_k$, a standard approach to generate pseudo-labels is greedy assignment, which assigns each sample to its nearest prototype based on feature similarity. However, this sample-wise independent strategy is highly vulnerable to feature bias caused by domain shift. It often over-assigns ambiguous samples to majority classes, resulting in severely biased pseudo-labels.

To overcome this limitation, we introduce the Global Distribution Alignment (GDA) module, which formulates pseudo-label generation as an Optimal Transport (OT) problem \cite{villani2008optimal}, as shown in Fig. \ref{fig:framework}. Unlike greedy assignment which optimizes locally, OT treats the assignment process as a global distribution alignment problem. It seeks a globally optimal alignment plan that minimizes the total transport cost while strictly enforcing specific marginal distribution constraints.
Crucially, this mechanism prevents majority classes from over-claiming ambiguous samples, ensuring that valid samples for minority classes are preserved.

\subsubsection{Global Cost Matrix Construction}
We formulate pseudo-label assignment as a minimum-cost transport problem between the sample feature set and the category prototype set. First, we construct a global affinity cost matrix $\mathbf{M} \in \mathbb{R}^{N \times K}$, where the element $M_{ik}$ measures the cosine distance between the $i$-th sample and the $k$-th prototype in the feature space:
\begin{equation}
M_{ik} = 1 - \frac{\langle G(x_i), \boldsymbol{\mu}_k \rangle}{\|G(x_i)\| \|\boldsymbol{\mu}_k\|}
\end{equation}
This matrix quantifies the geometric cost of assigning samples to respective categories.
The point-to-prototype formulation restricts the affinity matrix $\mathbf{M}$ to $N \times K$. Given that $K \ll N$, this reduces the computational complexity to linear $\mathcal{O}(N)$, ensuring scalability for massive point clouds.

\subsubsection{Entropy-Regularized Sinkhorn Optimization}
To achieve global distribution alignment, we solve for an optimal assignment matrix $\mathbf{Q}^*$, aiming to minimize the total transport cost while satisfying specific marginal distribution constraints. The optimization objective is defined as:
\begin{equation}
\mathbf{Q}^* = \mathop{\arg\min}_{\mathbf{Q} \in \mathbb{U}(\mathbf{r}, \mathbf{c})} \sum_{i=1}^N \sum_{k=1}^K Q_{ik} M_{ik} - \lambda H(\mathbf{Q})
\end{equation}
where $\lambda$ is the regularization coefficient, and $H(\mathbf{Q})$ is the entropy regularization term. The constraint set $\mathbb{U}(\mathbf{r}, \mathbf{c})$ formalizes the global prior:
\begin{equation}
\mathbb{U}(\mathbf{r}, \mathbf{c}) = \{ \mathbf{Q} \in \mathbb{R}_+^{N \times K} \mid \mathbf{Q}\mathbf{1}_K = \mathbf{r}, \mathbf{Q}^\top \mathbf{1}_N = \mathbf{c} \}
\end{equation}
Here, $\mathbf{r} = \frac{1}{N}\mathbf{1}_N$ enforces that every sample is assigned equal weight.
The vector $\mathbf{c} \in \mathbb{R}^K$ represents the global target class prior. In our offline setting, we assume that while individual predictions may contain noise, the aggregate statistics provide a stable estimate of the domain's class distribution. Thus, $\mathbf{c}$ is calculated from the global statistics of the ensemble predictions: $c_k = \frac{|\mathcal{I}_k|}{\sum_{j} |\mathcal{I}_j|}$. 
This transforms local statistical information into boundary conditions for global optimization.

We utilize the Sinkhorn-Knopp algorithm \cite{sinkhorn1967diagonal} to efficiently solve this convex optimization problem iteratively, ultimately obtaining the optimal assignment labels $\hat{y}_{sink, i} = \arg\max_k Q^*_{ik}$ from a global perspective. The Sinkhorn algorithm enforces an inter-class competition mechanism, compelling the model to seek an optimal solution that conforms to global class distribution.
In practice, executed via GPU-accelerated matrix normalizations with a strictly bounded number of iterations (e.g., only 3 iterations), this process is computationally lightweight and adds negligible overhead.

\subsection{Local-Global Dual-Consensus Filtering}

Even with robust prototypes and global alignment, individual pseudo-labels may still carry noise due to the inherent conflict between local feature similarity and global distributional constraints. Relying exclusively on either view can lead to biased supervision. To resolve this ambiguity, we propose the Local-Global Dual-Consensus Filtering (LGDCF) strategy, as shown in Fig. \ref{fig:framework}. The core insight is to exploit the complementarity of the two views: we identify the intersection between the multi-augmented ensemble prediction (Local View) and the Sinkhorn optimization assignment (Global View), retaining only those samples where feature-based confidence aligns with global distributional optimality. The final pseudo-label $\hat{y}_{final, i}$ is generated as:

\begin{equation}
\hat{y}_{final, i} = 
\begin{cases} 
\hat{y}_{raw, i}, & \text{if } \hat{y}_{raw, i} = \hat{y}_{sink, i} \\
\text{ignore}, & \text{otherwise}
\end{cases}
\end{equation}
This mechanism ensures reliability: When $\hat{y}_{raw} = \hat{y}_{sink}$, the sample is not only highly matched with a specific prototype in terms of local features but is also assigned to the same category after verification through global distribution constraints and inter-class competition. Such samples are regarded as reliable pseudo-labels. Conversely, when the two are inconsistent, the sample likely resides in an overlapping region with semantic ambiguity. In such cases, the local view might misjudge due to minor feature perturbations, or the global view might produce biased Sinkhorn assignments caused by potential statistical fluctuations in the estimated prior $\mathbf{c}$ under severe domain shifts. By flagging these conflicting samples as ignore and excluding them from gradient backpropagation, this strict intersection naturally guards against both local noise and global confirmation bias. It ensures that the model is supervised only by highly reliable pseudo-labels, driving a positive feedback loop. As the feature representations improve, the model progressively mitigates the domain shift, ultimately leading to robust and accurate segmentation on the target domain.

\subsection{Optimization and Temporal Ensembling}

Following the offline pseudo-label generation phase, we proceed to the parameter optimization stage. In this stage, the generated global pseudo-labels $\hat{y}_{final}$ are assigned to the target dataset $\mathcal{D}_T$. We then construct a data loader to iterate over $\mathcal{D}_T$ in mini-batches for standard supervised learning.

The student model $M_{stu}$ is trained by minimizing the Cross-Entropy loss on the filtered valid samples:
\begin{equation}
\mathcal{L}_{ce} = - \frac{1}{|\mathcal{B}_{valid}|} \sum_{x_i \in \mathcal{B}_{valid}} \log P(y = \hat{y}_{final, i} \mid x_i; \theta_{stu})
\end{equation}
where $\mathcal{B}_{valid} = \{x_i \in \mathcal{B} \mid \hat{y}_{final, i} \neq \text{ignore}\}$ denotes the set of valid samples within the current mini-batch $\mathcal{B}$.
By utilizing these high-quality pseudo-labels derived from the global consensus, the student model is guided to learn robust decision boundaries.

To further enhance the stability of the training process, the teacher model $M_{tea}$ does not participate in gradient descent but evolves through temporal ensembling of the student model parameters. Specifically, at each training step $t$, the teacher parameters $\theta_{tea}$ are updated as follows:
\begin{equation}
\theta_{tea}^{(t)} = \alpha \theta_{tea}^{(t-1)} + (1 - \alpha) \theta_{stu}^{(t)}
\end{equation}
where $\alpha$ is the momentum coefficient (set to 0.999 in our experiments). 
This EMA mechanism effectively smooths the optimization trajectory. Crucially, since the teacher model is used to generate pseudo-labels for the next epoch, this temporal smoothing ensures that the generated supervision signals remain stable and do not oscillate drastically between epochs, thereby preventing error amplification.

\section{Experiments}
\label{experiment}

\subsection{Dataset and Benchmark}

\subsubsection{Dataset Description}

STPLS3D \cite{Chen_2022_BMVC} is a large-scale aerial photogrammetry point cloud dataset containing both real-world and synthetic urban scenes, covering over $17 \text{ km}^2$. It comprises a $1.27 \text{ km}^2$ real-world subset reconstructed from UAV imagery, and a $16 \text{ km}^2$ synthetic subset generated by simulating UAV flight paths and reconstruction processes. The dataset features typical Bird's-eye View (BEV) geometry along with RGB attributes, and is annotated with 18 semantic categories.

H3D \cite{kolle2021hessigheim} is a high-resolution UAV LiDAR benchmark dataset featuring a dense point cloud with an average density of 800 pts/m$^2$. The labeled experimental subset comprises approximately 159 million points. Each point contains spatial coordinates $(X, Y, Z)$, reflectance, echo number, and RGB attributes. The dataset provides point-wise annotations for 11 semantic categories.

DALES \cite{varney2020dales} is a large-scale Airborne Laser Scanning (ALS) point cloud dataset covering a $10 \text{ km}^2$ area. Acquired from a high flight altitude of approximately $1300$ m, it contains over $505$ million points with a relatively sparse average density of about 50 pts/m$^2$. Each point includes spatial coordinates $(X, Y, Z)$, intensity, and return number attributes. The dataset is annotated with 8 semantic categories.

T3D \cite{tan2020toronto} is an urban MLS point cloud dataset covering a $1$ km road segment. It contains approximately $78.3$ million points and features a high point density of around 1000 pts/m$^2$. Point attributes include spatial coordinates $(X, Y, Z)$, RGB, intensity, GPS time, and scan angle. The dataset is annotated with 8 semantic categories.

\subsubsection{Benchmark}

\begin{figure*}[!htbp]
  \centering
   \includegraphics[scale=0.48]{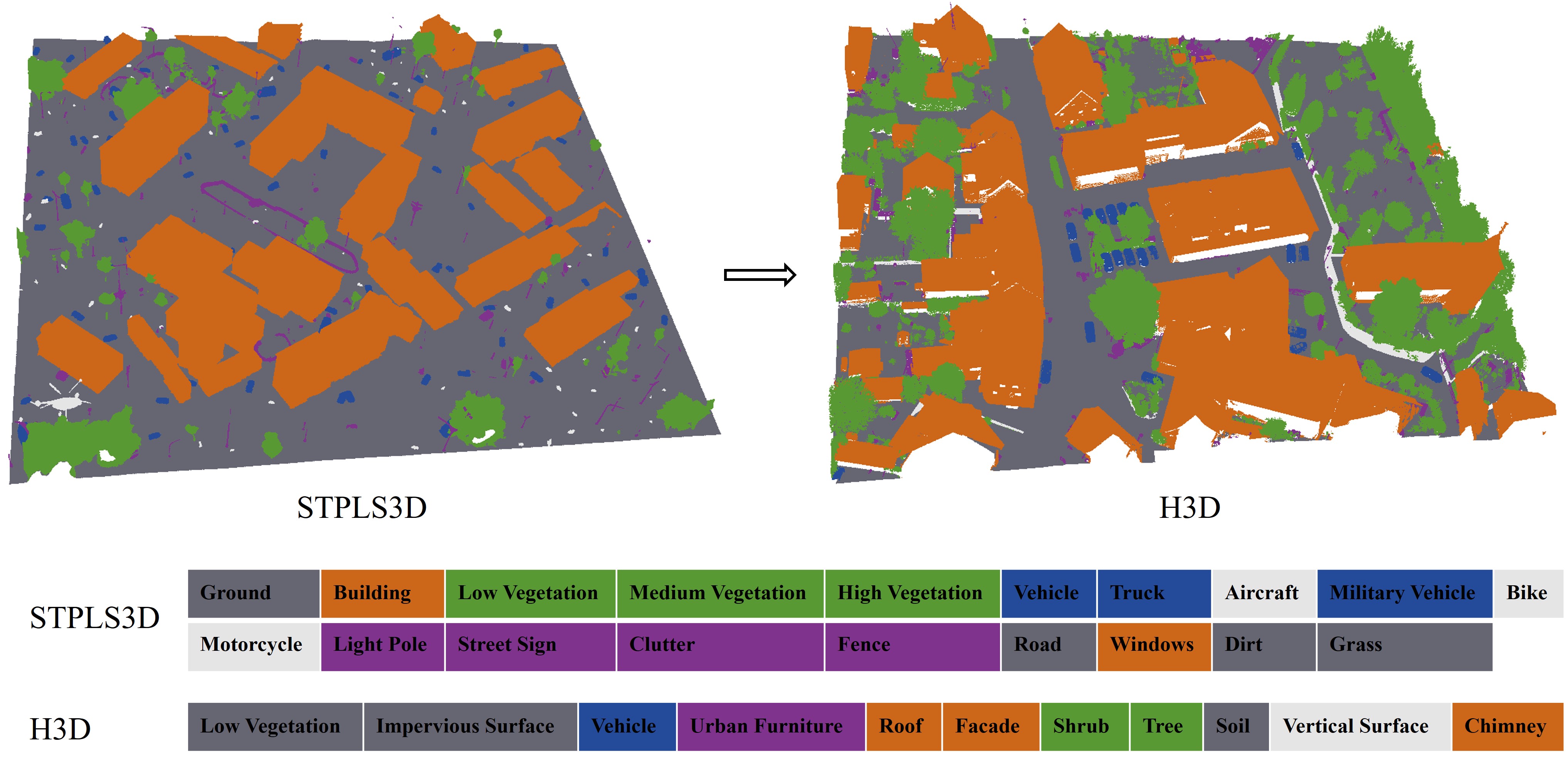}
   \caption{Visualization of the domain adaptation scenario from photogrammetry-derived point clouds (STPLS3D) to UAV-based LiDAR point clouds (H3D).  The top row compares the representative scenes, while the bottom row displays their corresponding semantic label spaces.}
  \label{fig:STPLS3DToH3D}
\end{figure*}

\begin{figure*}[!htbp]
  \centering
   \includegraphics[scale=0.51]{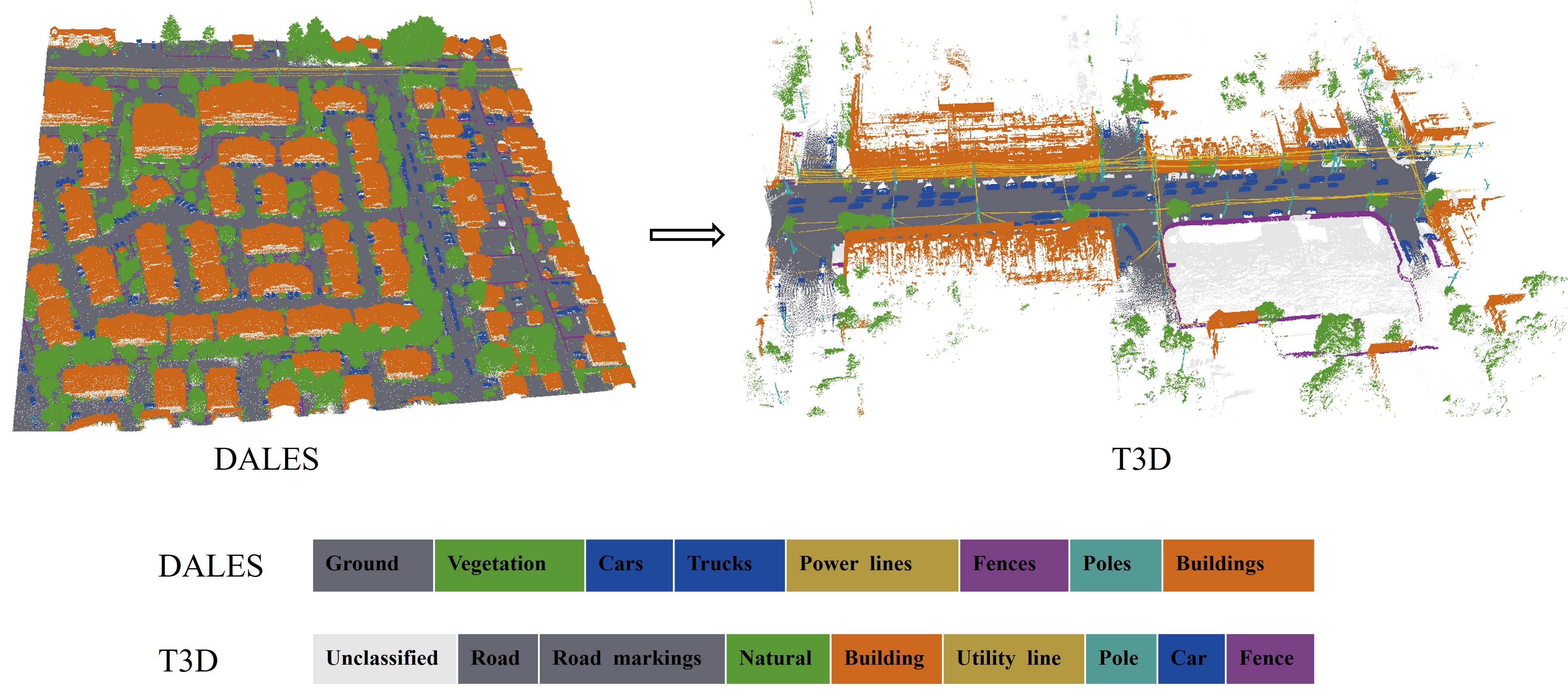}
  \caption{Visualization of the domain adaptation scenario from ALS point clouds (DALES) to MLS point clouds (T3D). The top row compares the representative scenes, while the bottom row displays their corresponding semantic label spaces.}
  \label{fig:DALESToT3D}
\end{figure*}

To evaluate our method in geospatial applications, we designed two transfer protocols addressing key domain shifts in urban mapping:

\textbf{Task 1: STPLS3D $\to$ H3D.}
This setting simulates a practical ``Low-Cost to High-Precision" transfer scenario, characterized by severe discrepancies in sensor modality and scene layout. The primary challenge stems from the fundamental structural gap between photogrammetry and LiDAR. The source data (STPLS3D), derived from passive aerial imagery, is prone to geometric noise and blurred structural boundaries, and critically, it lacks vegetation penetration capability, resulting in surface-only representations. In contrast, the target data (H3D) consists of precise, high-resolution UAV LiDAR scans (800 pts/m²) with sharp geometry and intensity attributes. This cross-sensor shift is further compounded by the scene divergence: transferring from synthetic environments generated by procedural pipelines to real-world German villages introduces significant discrepancies in architectural styles and object textures. We define 5 common categories for alignment: \textit{Ground}, \textit{Vehicle}, \textit{Urban Furniture}, \textit{Building}, and \textit{Vegetation} (Fig. \ref{fig:STPLS3DToH3D}).

\textbf{Task 2: DALES $\to$ T3D.}
This setting addresses the drastic ``Nadir-View to Street-View" shift, introducing extreme geometric and density variations driven by distinct acquisition mechanisms. The domain gap is profound: the source data (DALES) is collected via ALS from a high altitude of 1300 m, resulting in sparse point clouds (50 pts/m²) that predominantly capture rooftops and large-scale terrain. Conversely, the target data (Toronto-3D) is acquired by MLS along a single road segment, offering dense street-level details but missing building tops. Furthermore, the scene context shifts from wide-area aerial surveys to focused urban street corridors. The model must essentially learn to infer detailed street-level semantics from sparse top-down priors. We define 7 common categories for alignment: \textit{Road}, \textit{Natural}, \textit{Car}, \textit{Utility line}, \textit{Fence}, \textit{Pole}, and \textit{Building} (Fig. \ref{fig:DALESToT3D}).

\subsection{Implementation Details}

All models are implemented in PyTorch 1.8.2 on a single NVIDIA GeForce GTX 3090 24GB GPU. We adopt KPConv (KPFCNN) \cite{thomas2019kpconv} as the 3D backbone, following the default configuration of the official open-source repository.
We rely solely on 3D coordinates to ensure the method's generalizability across sensors with diverse attribute configurations.
To accommodate the density and scale differences across datasets, we apply different grid subsampling sizes $dl_0$ and input sphere radii $R_{\text{in}}$: for STPLS3D, $dl_0 = 0.3 \text{m}$, $R_{\text{in}} = 30 \text{m}$; for DALES, $dl_0 = 0.2 \text{m}$, $R_{\text{in}} = 15 \text{m}$.

In the adaptation stage, we freeze all weights except for the BN affine parameters. The student model is updated using the SGD optimizer with a reduced learning rate of $1 \times 10^{-3}$. The teacher model evolves via EMA with a momentum of $\alpha = 0.999$. For H3D, we set the number of steps per epoch to 50 with a batch size of 2. For the T3D dataset, the steps are set to 40 with a batch size of 8. Additionally, for the H3D dataset, the ensemble size in the stochastic multi-augmented ensemble is set to $V = 4$, and the anchor selection ratio for class-balanced local prototype estimation is set to $\rho = 0.8$. 
Empirically, for the more challenging T3D dataset, we set the anchor selection ratio to $\rho = 0.7$.

\subsection{Evaluation Metrics}

We employ mean Intersection-over-Union (mIoU) and Overall Accuracy (OA) to assess the performance of our method. Let $K$ denote the total number of semantic classes. For each class $k \in \{1, \dots, K\}$, we calculate the number of True Positives ($\text{TP}_k$), False Positives ($\text{FP}_k$), and False Negatives ($\text{FN}_k$) based on the confusion matrix. The Intersection-over-Union ($\text{IoU}_k$) for class $k$ is computed as the ratio of the intersection to the union between the ground-truth and the prediction:
\begin{equation}
\text{IoU}_k = \frac{\text{TP}_k}{\text{TP}_k + \text{FP}_k + \text{FN}_k}
\end{equation}
The final mIoU is the macro-average of the IoU scores over all $K$ classes:
\begin{equation}
\text{mIoU} = \frac{1}{K} \sum_{k=1}^{K} \text{IoU}_k
\end{equation}

OA measures the global classification accuracy. It is calculated as the ratio of the total number of correctly classified points to the total number of points:
\begin{equation}
\text{OA} = \frac{\sum_{k=1}^{K} \text{TP}_k}{N}
\end{equation}
where $\sum_{k=1}^{K} \text{TP}_k$ represents the total correct predictions across all categories. OA serves as a supplementary metric to reflect global consistency.

\subsection{Experiment Results}

\begin{figure*}[htbp]
  \centering
   \includegraphics[scale=0.48]{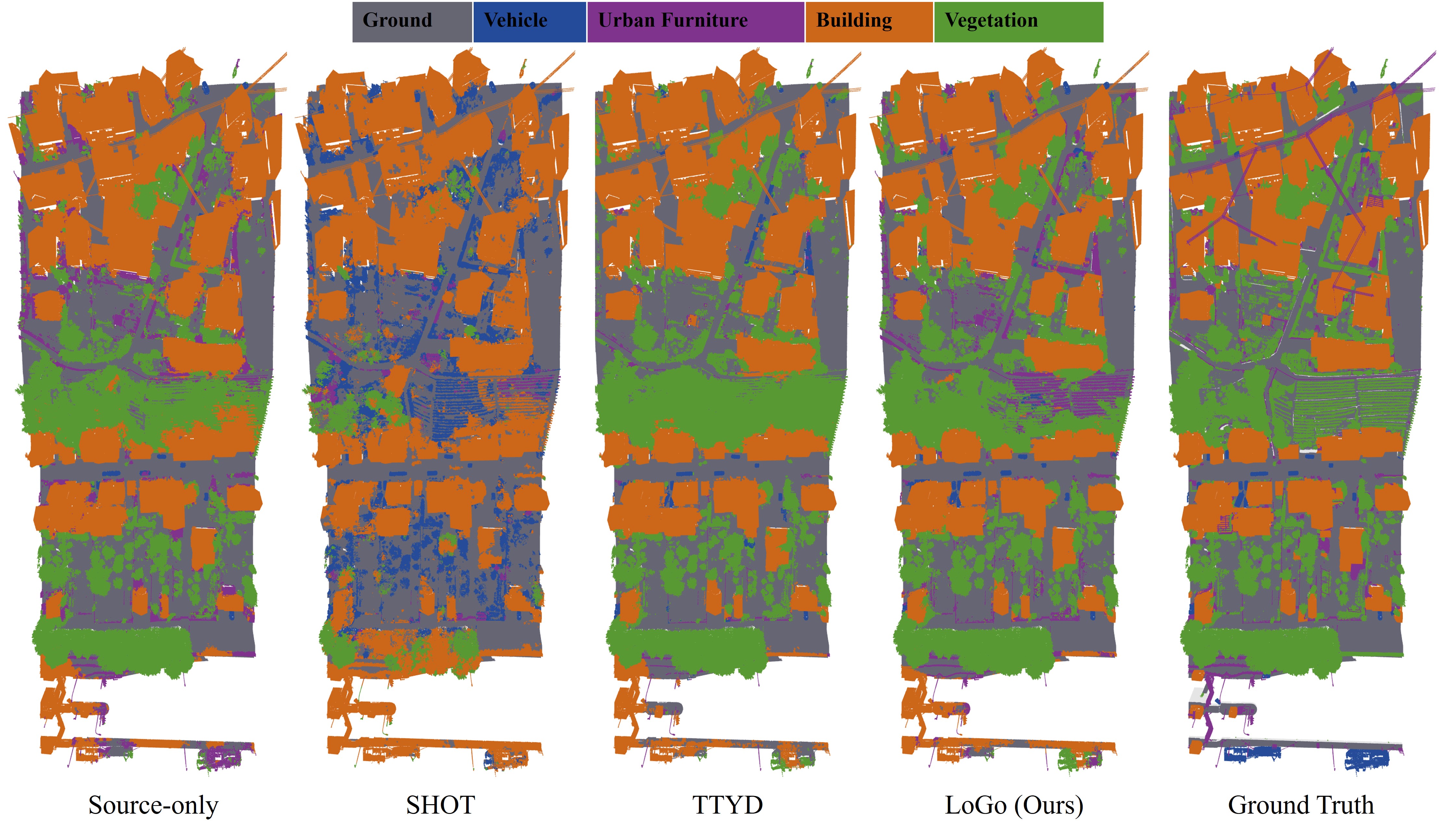}
  \caption{
    Qualitative comparison of global semantic segmentation results on the H3D dataset. From left to right: Source-only, SHOT, TTYD, LoGo, and Ground Truth. 
  }
  \label{fig:H3D_Result}
\end{figure*}

To validate the effectiveness of our method, we compare it against several representative approaches:

\textbf{Source-only}: This baseline directly applies the model pre-trained on the source domain to the target domain for inference without any adaptation.

\textbf{AdaBN} \cite{li2016revisiting}: This method replaces the Batch Normalization (BN) layer statistics (mean and variance) of the source model with those estimated from the target domain data during inference, achieving adaptation without updating any model parameters.

\textbf{TENT} \cite{wang2021tent}: This method leverages unlabeled target data during inference. By minimizing the entropy of model predictions, it updates only the BN layer statistics and affine parameters, adapting to the target domain distribution shift.

\textbf{SHOT} \cite{liang2020we}: SHOT freezes the source classifier and fine-tunes only the feature extraction module. It incorporates the Information Maximization (IM) principle to encourage the model to produce discriminative and globally diverse outputs. Furthermore, it employs a self-supervised pseudo-labeling strategy by computing class centroids within the target feature space to generate reliable pseudo-labels, thereby guiding feature alignment.
     
\textbf{SHOT-ELR} \cite{yi2023when}: Addressing potential pseudo-label noise, this method builds upon SHOT by introducing a plug-and-play Early Learning Regularization (ELR) term. It utilizes the temporal moving average of historical predictions to guide current training, aiming to prevent overfitting to erroneous pseudo-labels.

\textbf{TTYD} \cite{michele2024train}:  This method fine-tunes the BN layers of the source model by minimizing prediction entropy and using source domain priors. It calculates prediction consistency between the current and a reference model (e.g., a PTBN model with adapted statistics only) to detect model degradation and trigger early stopping. This metric is also used for hyperparameter selection on the unlabeled target domain. An optimal model is then selected for pseudo-label generation in a second self-training stage.
     
\textbf{Wang et al.} \cite{WANG2025422}: This method fine-tunes only the BN layers of the pre-trained model. It employs a Progressive Batch Normalization (PBN) module to continuously incorporate test data statistics into the model. A self-supervised optimization strategy encourages high-confidence predictions through information maximization. Reliable pseudo-labels are generated by jointly evaluating entropy-based confidence and contrastive consistency.

\begin{table*}[!htbp]
\caption{Quantitative comparison with state-of-the-art methods on the STPLS3D $\to$ H3D transfer task. The reported metrics include per-class IoU, mean IoU (mIoU), and Overall Accuracy (OA). The best results are highlighted in \textbf{bold}, and the second-best results are \underline{underlined}.}
\label{tab:h3d}
\centering
\small
\begin{tabular}{lccccccccc}
\toprule
\multirow{2}{*}{Method} & \multicolumn{5}{c}{IoU} & \multirow{2}{*}{mIoU} & \multirow{2}{*}{OA} \\ 
\cmidrule(lr){2-6}  & Ground &  Vehicle & Urban Fur. & Building & Vegetation  \\ \midrule                        
Source-only   & 76.73 & 20.01 & \underline{12.22} & 79.49 & 55.71 & 48.83 & 82.12 \\
AdaBN \cite{li2016revisiting} & 77.69 & 28.17 & 10.01 & 80.02 & 54.46 & 50.07 & 82.86 \\
TENT \cite{wang2021tent}    & 77.46 & 23.96 & 9.93  & 79.92 & 56.42 & 49.54 & 83.07 \\
SHOT \cite{liang2020we}    & 78.91 & 4.40  & 2.48  & 66.22 & 13.97 & 33.20 & 72.80 \\
SHOT-ELR \cite{yi2023when} & \textbf{84.83} & 20.01 & 11.31 & \underline{86.33} & 49.67 & 50.43 & \underline{84.67} \\
TTYD \cite{michele2024train}   & 76.06 & \textbf{33.82} & 6.56  & 85.74 & \underline{56.45} & \underline{51.73} & 84.07 \\ \midrule
LoGo (Ours)    & \underline{81.47} & \underline{30.84} & \textbf{13.10} & \textbf{87.61} & \textbf{60.74} & \textbf{54.75} & \textbf{85.98} \\
\bottomrule
\end{tabular}
\end{table*}

\subsubsection{Analysis on STPLS3D $\to$ H3D}

Table \ref{tab:h3d} details the transfer results from synthetic photogrammetric point clouds to real-world UAV LiDAR data. This cross-modality setting presents a challenging validation scenario due to the fundamental differences in imaging mechanisms (passive optical reconstruction and active laser scanning) and distinct noise patterns between the source and target domains.

\begin{figure*}[t]
  \centering
   \includegraphics[scale=0.5]{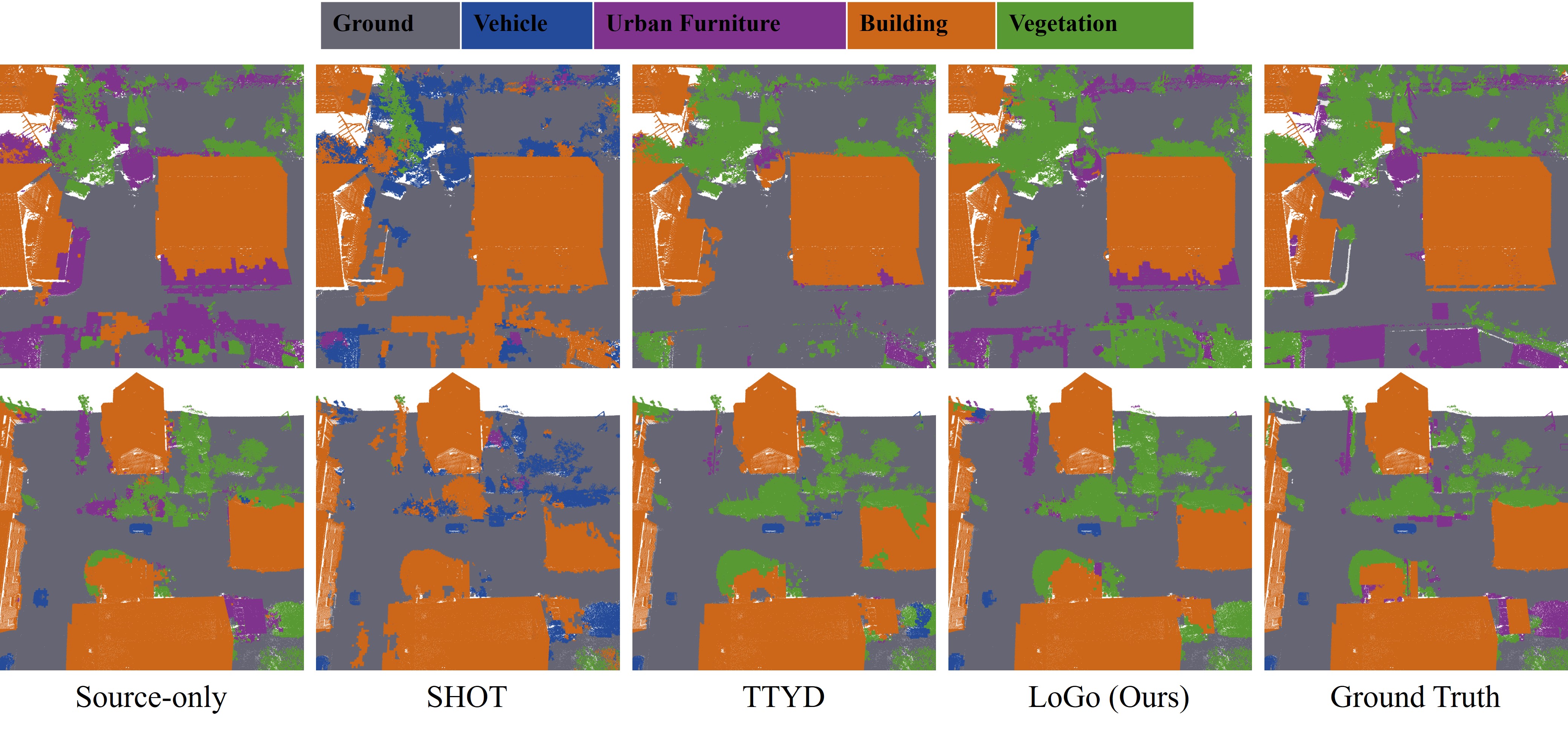}
  \caption{
        Qualitative comparison of local semantic segmentation details on the H3D dataset.
  }
  \label{fig:H3D_Result2}
\end{figure*}

The Source-only baseline exhibits poor generalization. While majority classes (e.g., Ground, Building) maintain reasonable performance, Vehicle samples are frequently misclassified as Urban Furniture (see Fig. \ref{fig:H3D_Result}). Lightweight adaptation methods (AdaBN, TENT) achieve only marginal overall improvements. Although they slightly improve the Vehicle category, they suffer from degradation in tail classes. This suggests that solely minimizing prediction entropy on noisy target data exacerbates ``confirmation bias," compelling the model to become over-confident in erroneous predictions.

Notably, SHOT exhibits severe ``negative transfer," with its mIoU dropping to 33.20\% (a 15.63\% decrease compared to Source-only). As depicted in Fig. \ref{fig:H3D_Result}, numerous samples are misidentified as Vehicle. We attribute this failure to SHOT's greedy distance-based matching. Given the highly distorted target features, this Euclidean distance-based matching triggers a ``Winner-Takes-All" phenomenon, where ambiguous samples erroneously gravitate towards dominant class centroids.
SHOT-ELR attempts to improve robustness by regularizing training with a moving average of historical predictions, but its improvement is limited. TTYD attempts adaptation through entropy minimization and source domain priors, but this can lead to accuracy degradation in small categories like Urban Furniture, misidentifying them as Building, as shown in Fig. \ref{fig:H3D_Result2}.

In contrast, our LoGo framework achieves the best performance across all metrics (mIoU 54.75\%, OA 85.98\%), surpassing the Source-only baseline by 5.92\% and avoiding the catastrophic failure seen in SHOT. This success validates our synergistic design. By leveraging CBLPE, we successfully preserve feature anchors for sparse categories (e.g., Urban Furniture). Crucially, the GDA module suppresses the ``Winner-Takes-All" bias by enforcing global Optimal Transport constraints, which prevents dominant classes from over-claiming ambiguous samples and regularizes the assignment process. Finally, the LGDCF dual-consensus strategy strictly filters out inconsistent pseudo-labels. Consequently, LoGo achieves an impressive 13.10\% IoU on the challenging Urban Furniture category (substantially outperforming TTYD) and maintains a robust 60.74\% IoU on Vegetation. These results demonstrate the effectiveness of LoGo in balancing local feature learning and global distributional constraints, successfully mitigating the severe domain shifts caused by heterogeneous sensors.

\subsubsection{Analysis on DALES $\to$ Toronto-3D}

Table \ref{tab:dales_t3d} details the transfer results from the ALS dataset to the MLS dataset. This setting involves a significant viewpoint shift from a ``high-altitude nadir view" to a ``ground-level street view," leading to distinct performance among different adaptation mechanisms.

\begin{table*}[!htbp]
\caption{Quantitative comparison with different methods on the DALES $\to$ T3D transfer task. The reported metrics include per-class IoU, mean IoU (mIoU), and Overall Accuracy (OA). The best results are highlighted in \textbf{bold}, and the second-best results are \underline{underlined}. Note that the results for Wang et al. \cite{WANG2025422} are cited directly from the original paper.}
\label{tab:dales_t3d}
\centering
\small
\begin{tabular}{lccccccccc}
\toprule
\multirow{2}{*}{Method} & \multicolumn{7}{c}{IoU} & \multirow{2}{*}{mIoU} & \multirow{2}{*}{OA} \\ 
\cmidrule(lr){2-8}  & Road & Natural & Car & Utility line & Fence & Pole & Building  \\        \midrule                        
Source-only   & \underline{97.96} & 77.84 & 46.89 & 65.35 & 5.81 & 50.53 & 48.27 & 56.09 & 92.29 \\
AdaBN \cite{li2016revisiting}  & 97.42 & 67.06 & 12.50 & 51.34 & 7.27 & 15.57 & 31.35 & 40.36 & 88.82 \\
TENT \cite{wang2021tent}     & 97.46 & 63.25 & 8.78  & 47.63 & 6.81 & 12.12 & 19.64 & 36.53 & 87.45 \\
SHOT \cite{liang2020we}   & 97.17 & 85.97 & 62.00 & \underline{65.80} & \underline{16.40} & 30.70 & \underline{71.14} & 61.31 & \underline{94.34} \\
SHOT-ELR \cite{yi2023when} & 96.73 & 60.86 & 66.71 & 0.00     & 0.16 & 0.00  & 1.30  & 32.25 & 86.75 \\
TTYD \cite{michele2024train} & 97.35 & 72.58 & 0.43  & 52.04 & 7.50 & 10.02 & 50.79 & 41.53 & 90.33 \\
Wang et al. \cite{WANG2025422} & \textbf{98.09} & \underline{86.99} & \underline{72.72} & 60.12 & 3.07 & \underline{60.34} & 53.50 & \underline{62.12} & 92.67 \\
\midrule
LoGo (Ours)& 97.85 & \textbf{92.30} & \textbf{83.50} & \textbf{72.95} & \textbf{21.13} & \textbf{62.59} & \textbf{84.46} & \textbf{73.54} & \textbf{96.93} \\
\bottomrule
\end{tabular}

\end{table*}

\begin{figure*}[b]
  \centering
   \includegraphics[scale=0.44]{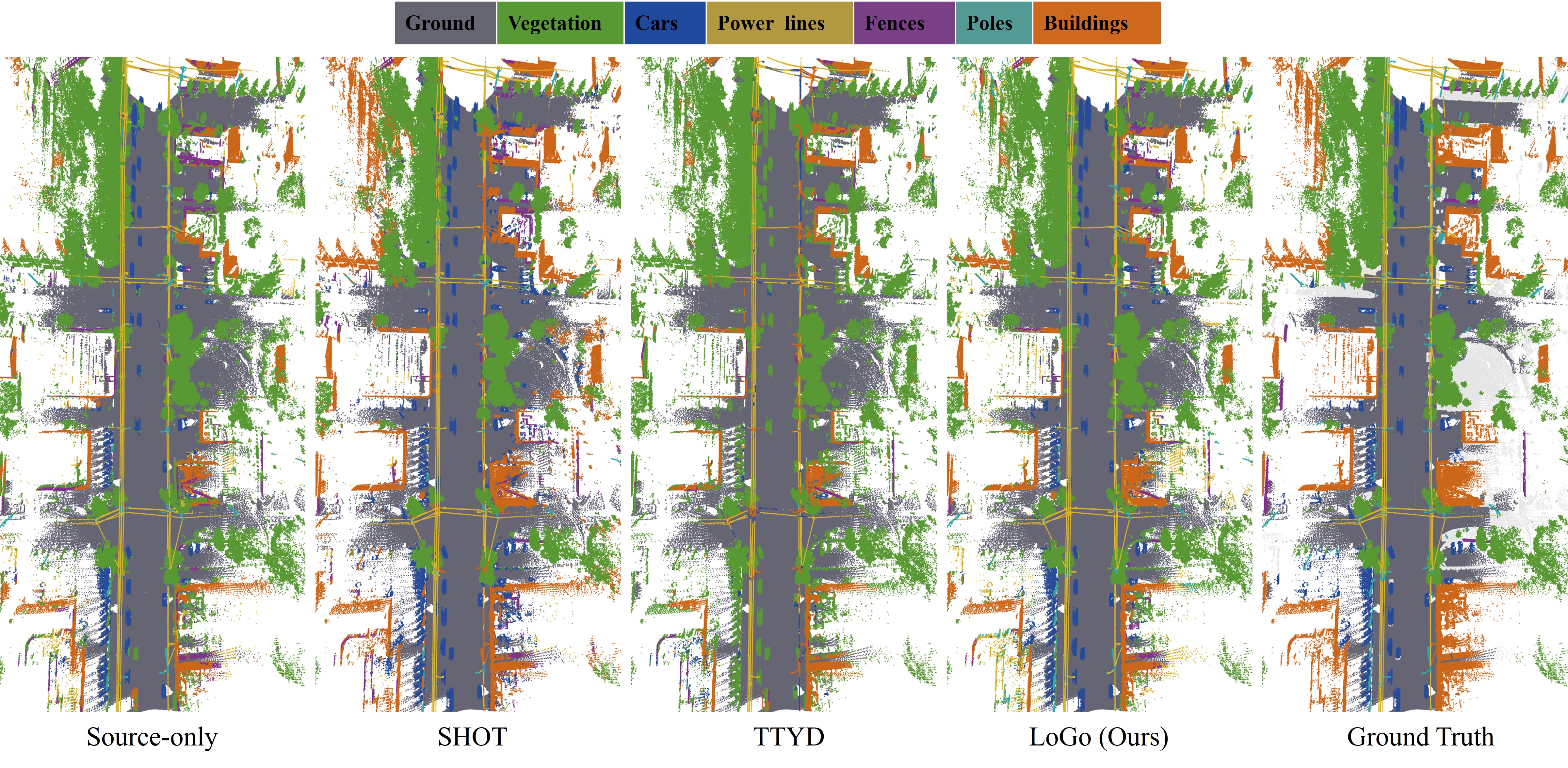}
  \caption{
 Qualitative comparison of global semantic segmentation results on the T3D dataset. From left to right: Source-only, SHOT, TTYD, LoGo, and Ground Truth. 
  }
  \label{fig:T3D_Result}
\end{figure*}

The AdaBN baseline, which replaces BN statistics with local target domain means, drops the mIoU to 40.36\%. This is due to the spatially non-uniform scanning of MLS data, which skews local statistics and compromises decision boundaries. TENT further exacerbates this decline to 36.53\%. Its strategy of minimizing local batch entropy leads to overfitting on dominant classes (e.g., Road) while triggering the forgetting of sparse categories like Fence.

\begin{figure*}[!htbp]
  \centering
   \includegraphics[scale=0.41]{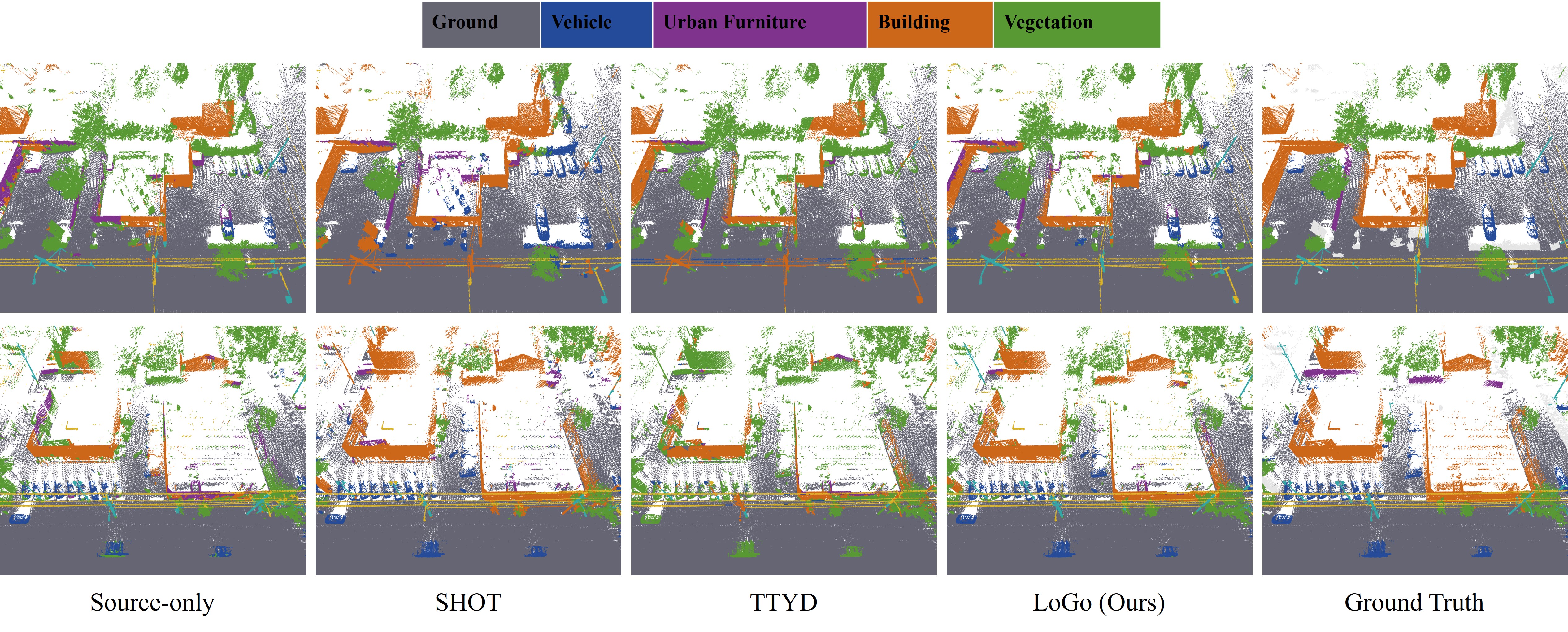}
  \caption{
     Qualitative comparison of local semantic segmentation details on the T3D dataset.
  }
  \label{fig:T3D_Result2}
\end{figure*}

SHOT utilizes global prototype estimation to mitigate local noise, raising mIoU to 61.31\%. 
However, its distance-based greedy matching can cause misclassifications, such as Pole being misclassified as Building (Fig. \ref{fig:T3D_Result2}). In contrast, SHOT-ELR suffers a performance drop (mIoU 32.25\%) because its ELR mechanism proves overly conservative. It erroneously identifies key geometric features distorted by viewpoint shifts (e.g., Utility line) as label noise and suppresses them, hindering the model's ability to learn hard samples. TTYD is severely limited by ``Label Distribution Shift." By forcing the prediction distribution of MLS data to align with the discrepant ALS prior, it causes the target-specific dominant category (e.g., Car) to be erroneously under-represented, with accuracy dropping to 0.43\% (Fig. \ref{fig:T3D_Result}).
Additionally, the method of Wang et al. achieves the best result on the Road category (98.09\%), benefiting from its progressive BN update strategy's ability to smoothly adapt to spatially dominant categories with stable feature statistics. However, its failure on Fence (3.07\%) again exposes the shortcoming of online TTA. When facing long-tailed distributions, the scarcity of rare categories in local batches prevents the accumulation of sufficient gradients, ultimately leading to the ``Winner-Takes-All" effect.

In contrast, our method achieves a mIoU of 73.54\%, demonstrating notable robustness across all metrics. This success directly validates our algorithmic design. By utilizing CBLPE, we successfully preserve feature anchors for sparse classes, significantly alleviating forgetting and boosting the Fence IoU to 21.13\%. 
Furthermore, the GDA module rectifies the ``Winner-Takes-All" bias inherent in SHOT's greedy strategy, while simultaneously avoiding the mismatched label distribution constraints imposed by TTYD's source priors. By aligning samples based on dynamically estimated target priors, GDA ensures structurally consistent predictions for geometric objects like Utility lines and Poles. Finally, the LGDCF module rigorously filters samples through a ``local-global" dual verification mechanism. This effectively mitigates the propagation of pseudo-label noise, preventing the accumulation of erroneous supervision signals and enhancing adaptation stability in complex scenes.

\subsection{Experimental Analysis}
\subsubsection{Ablation Study}

To evaluate the efficacy of each module, we compare the Source-only baseline with two variants of our framework. Since our CBLPE module is strictly a representation-learning component designed to estimate target-domain class prototypes without performing specific label assignment tasks, it cannot be ablated independently. Therefore, we evaluate the following configurations:
(1) GDA: This variant introduces the global distribution alignment mechanism. Under this setting, instead of relying on local matching, pseudo-labels are derived directly from the global assignment plan computed by the Optimal Transport mechanism, thereby enforcing distribution-level constraints.
(2) LoGo (Full Method): This setting activates the complete Local-Global Dual-Consensus Filtering mechanism. Under this mode, the final supervision signal is determined by taking the intersection of the initial ensemble predictions and the global transport assignments, retaining only those samples where both perspectives reach a consensus.

Analysis. As shown in Table \ref{tbl:ablation}, the ablation results underscore the necessity of our holistic Local-Global framework. The Source-only baseline (Row 1) shows limited performance due to significant domain shift. 
The integration of GDA (Row 2) leads to a substantial improvement over the baseline on both datasets. This demonstrates that introducing global distribution constraints and inter-class competition effectively rectifies the assignment biases inherent in the local view. 
Ultimately, the Full Method (Row 3) achieves the optimal performance by enforcing the dual-consensus mechanism. By rigorously filtering out ambiguous boundary samples where the local and global views conflict, this strategy yields reliable pseudo-labels, effectively mitigating noise propagation as evidenced by the notable 73.54\% mIoU on the T3D dataset.

\begin{table}[t]
    \centering
    \caption{Ablation study of the proposed framework on STPLS3D $\to$ H3D and DALES $\to$ T3D. The best results are highlighted in \textbf{bold}.}
    \begin{tabular}{cccccc}
    \toprule
        \multicolumn{2}{c}{Module} & \multicolumn{2}{c}{STPLS3D $\to$ H3D} &  \multicolumn{2}{c}{DALES $\to$ T3D} \\ \cmidrule(lr){1-2} \cmidrule(lr){3-4} \cmidrule(lr){5-6}
        GDA & LGDCF & mIoU & OA & mIoU & OA \\ \midrule
        ~ & ~ & 48.83 & 82.12 & 56.09 & 92.29 \\ 
        \checkmark & ~ & 52.75 & 85.02 & 66.14 & 95.19 \\ 
        \checkmark & \checkmark & \textbf{54.75} & \textbf{85.98} & \textbf{73.54} & \textbf{96.93} \\ \bottomrule
    \end{tabular}
    \label{tbl:ablation}
\end{table}

\subsubsection{Parameter Sensitivity Analysis}
\label{sec:sensitivity}

\begin{figure}[!htbp]
  \centering
   \includegraphics[scale=0.4]{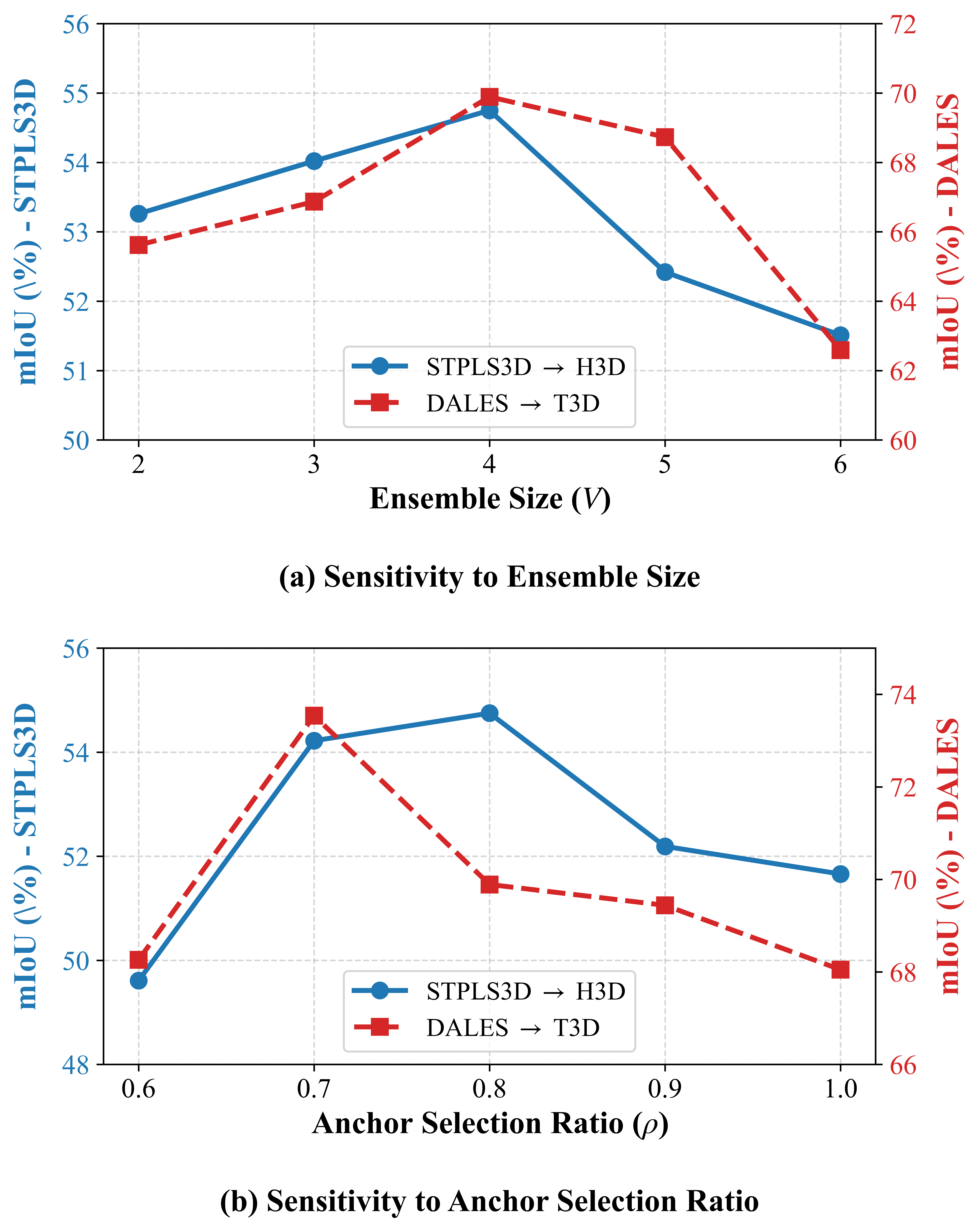}
  \caption{
    Parameter sensitivity analysis on STPLS3D $\to$ H3D and DALES $\to$ T3D tasks. (a) Impact of the ensemble size $V$ on segmentation performance; (b) Impact of the anchor selection ratio $\rho$ on local prototype estimation.
  }
  \label{fig:sensitivity}
\end{figure}

To evaluate the robustness of our proposed framework and the influence of key hyperparameters, we conduct a detailed sensitivity analysis on the STPLS3D $\to$ H3D and DALES $\to$ T3D benchmarks. We focus on two core parameters: the \textit{Ensemble Size} ($V$) used in the multi-augmented inference, and the \textit{Anchor Selection Ratio} ($\rho$) used in the class-balanced local prototype estimation. The quantitative results are illustrated in Figure \ref{fig:sensitivity}.

\textbf{Ensemble Size} $V$.
As shown in Figure \ref{fig:sensitivity} (a), segmentation performance increases initially before decreasing as $V$ grows. We observe a significant improvement in mIoU on both datasets when $V$ increases from 2 to 4. This indicates that a moderate ensemble size effectively exploits the prediction consistency across different geometric transformations, thereby smoothing out prediction variance during inference. However, when $V$ exceeds 4 (e.g., $V=6$), the performance suffers a slight decline. We attribute this to the fact that excessive stochastic augmentations might introduce distortions, causing predictions in certain views to deviate from the true distribution and degrading the quality of pseudo-labels. Consequently, we adopt $V=4$ as an optimal balance between adaptation accuracy and computational cost.

\textbf{Anchor Selection Ratio $\rho$.}
The parameter $\rho$ determines the proportion of high-confidence samples retained for calculating class prototypes. As depicted in Figure \ref{fig:sensitivity} (b), the model maintains highly competitive performance across a broad range, consistently outperforming existing state-of-the-art methods on both datasets. This underscores the critical trade-off between representativeness and noise robustness. Selecting the top 70\%-90\% of samples ensures that the computed prototypes are sufficiently representative of the intra-class distribution while effectively filtering out ambiguous samples near decision boundaries. Notably, setting $\rho=1.0$ (i.e., using all predicted samples) leads to a performance drop, confirming the necessity of our Top-K anchor mining strategy in preventing prototype contamination by noisy pseudo-labels. More importantly, even under this extreme setting without any filtering ($\rho=1.0$), LoGo still achieves an mIoU of over 68\% on the challenging DALES $\to$ T3D task, which comfortably surpasses the current state-of-the-art baseline (62.12\%). 
This compellingly demonstrates the inherent superiority and robustness of our fundamental Local-Global architectural design.

\section{Conclusion}
\label{conclus}
In this paper, we address the challenging problem of source-free unsupervised domain adaptation for semantic segmentation of heterogeneous geospatial point clouds under privacy and regulatory constraints. To this end, we propose LoGo, a self-training framework that effectively mitigates severe cross-scene and cross-sensor domain shifts without access to source data. Extensive experiments on challenging cross-domain benchmarks demonstrate consistent and significant performance gains. These results establish SFUDA as a crucial and feasible paradigm for large-scale geospatial point cloud semantic segmentation in real-world surveying and remote sensing scenarios, and demonstrate that robust adaptation critically depends on the synergy between local class-balanced feature discrimination and global distributional alignment when learning from unlabeled and long-tailed point clouds. Future work will focus on improving the robustness of domain adaptation under more extreme label distribution shifts, and extending the SFUDA framework to handle target scenes with missing or unseen categories.

\bibliographystyle{IEEEtran}
\bibliography{ref}

\vfill

\end{document}